\documentclass[final]{article}

\usepackage{mendel}

\usepackage[utf8]{inputenc} 
\usepackage[T1]{fontenc}    
\usepackage{hyperref}       
\usepackage{url}            
\usepackage{booktabs} 
\usepackage{amsmath}
\usepackage{amsfonts}       
\usepackage{nicefrac}       
\usepackage{microtype}      
\usepackage{xcolor}         
\usepackage{natbib}
\usepackage{graphicx}
\usepackage{multicol}
\usepackage{multirow}
\usepackage{float}

\title{ACR: A Benchmark for Automatic Cohort Retrieval}

\author{
$\text{Dung Ngoc Thai}^1\dagger$\And
$\text{Victor Ardulov}^1\dagger$\And 
$\text{Jose Ulises Mena}^1$\And
$\text{Simran Tiwari}^1$\And
$\text{Gleb Erofeev}^1$\And
$\text{Ramy Eskander}^2$\thanks{Work done while at Mendel AI} \And
$\text{Karim Tarabishy}^1$\And
$\text{Ravi B Parikh}^3$ \And
$\text{Wael Salloum}^1\dagger$\AND
\\
$\text{Mendel AI}^1$\\
$\text{Columbia University}^2$\\
$\text{University of Pennsylvania}^3$\\
\texttt{\{dung.t, victor.a, wael\}@mendel.ai}
}

\begin{document}
\newcommand{\nousecase}{6}
\newcommand{\checklist}{\textsc{ClinCheckList}}
\newcommand{\nochecklist}{100}
\newcommand{\nopatient}{1436}
\newcommand{\nodocument}{115,865}
\newcommand{\nodocumentshort}{116K}
\newcommand{\maxnumdoc}{1,251}
\newcommand{\avgnumdoc}{81}
\newcommand{\nocells}{162,268} 
\newcommand{\nosite}{4}
\newcommand{\novalquery}{20}
\newcommand{\novalpatient}{50}
\newcommand{\noquery}{113}
\newcommand{\kappascore}{1}
\maketitle

\begin{abstract}
Identifying patient cohorts is fundamental to numerous healthcare tasks, including clinical trial recruitment and retrospective studies. Current cohort retrieval methods in healthcare organizations rely on automated queries of structured data combined with manual curation, which are time-consuming, labor-intensive, and often yield low-quality results. Recent advancements in large language models (LLMs) and information retrieval (IR) offer promising avenues to revolutionize these systems. Major challenges include managing extensive eligibility criteria and handling the longitudinal nature of unstructured Electronic Medical Records (EMRs) while ensuring that the solution remains cost-effective for real-world application. This paper introduces a new task, Automatic Cohort Retrieval (ACR), and evaluates the performance of LLMs and commercial, domain-specific neuro-symbolic approaches. We provide a benchmark task, a query dataset, an EMR dataset, and an evaluation framework. Our findings underscore the necessity for efficient, high-quality ACR systems capable of longitudinal reasoning across extensive patient databases.
\end{abstract}

\section{Introduction}
\label{sec:intro}

\begin{figure}[t!]
\vspace{1em}
  \centering
  \includegraphics[height=0.5\textheight]{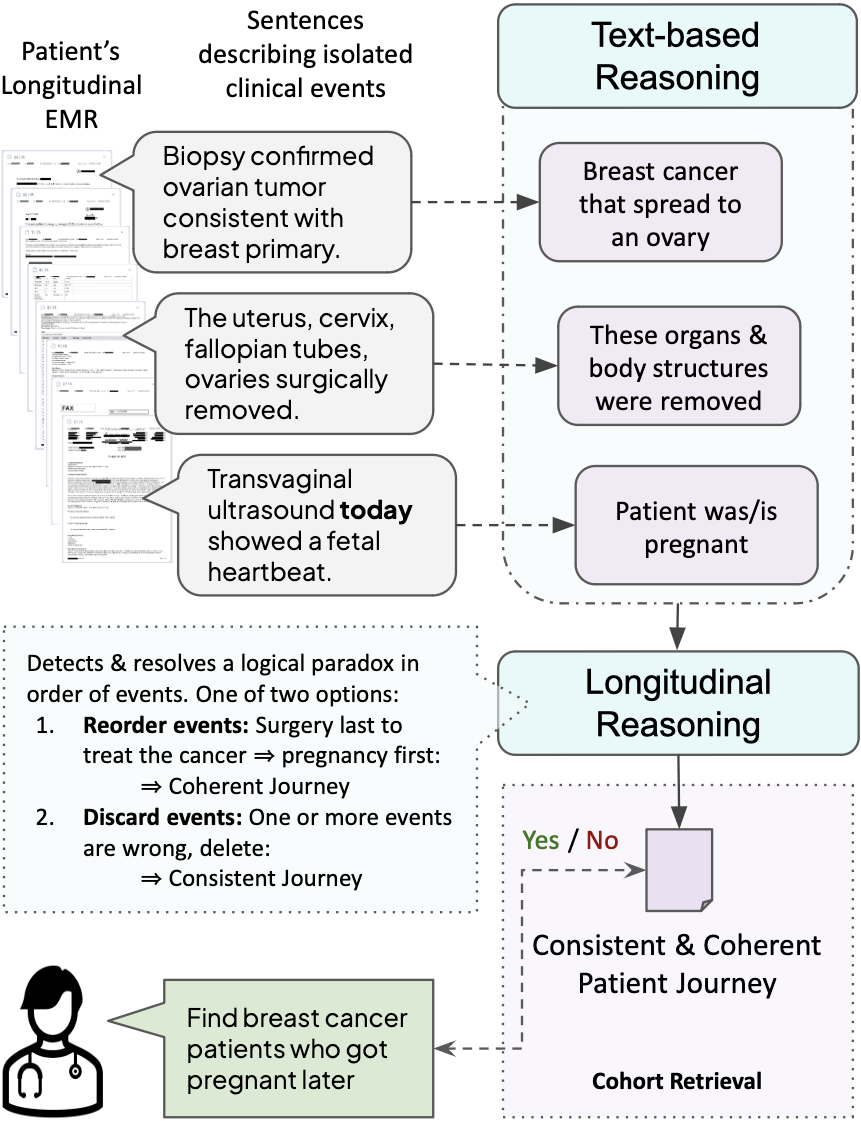} 
  \caption{\textbf{Longitudinal data challenges in Cohort Retrieval:} 
  This example shows a patient medical journey 
    depicting 3 related facts
    scattered in 3 documents written over the years.  
  Cohort retrieval systems must possess \textit{longitudinal reasoning} capacities to accurately answer the user query shown above.}
  \label{fig:longitudinal}
\end{figure}



Cohort retrieval 
    is a foundational task in healthcare 
    facilitating a wide range of applications in clinical research and practice, 
    including clinical trial recruitment and feasibility assessment, protocol design, and retrospective studies.
This process involves identifying a group, or cohort, of patients 
    from a real-world data (RWD) repository
    based on an inquiry defining patient eligibility criteria. 
Current standard of care for cohort retrieval relies primarily on automated queries of limited structured data combined with manual human curation, which can take hours for tasks like retrospective cohort studies, registry creation, or clinical trial prescreening \citep{ni2019real,vassar2013retrospective}.

One of the challenges for cohort retrieval 
    is the extensive eligibility criteria 
        that often grow to dozens of inclusion and exclusion criteria 
            covering various aspects of a patient's clinical history
    such as patient demographics, detailed characteristics of diseases and conditions,  treatment history, and specific lab measurements.
Another challenge is the complex and siloed nature of real-world data, 
    which refers to a wide range of patient-related data such as
    structured and unstructured Electronic Medical Records (EMRs), pharmacy records, insurance claims, patient registries, and even web, social media, and wearable device data.

In this work, we limit our RWD scope to unstructured EMRs for two main reasons: 
    they represent the richest source of RWD 
    and pose a significant challenge for cohort retrieval.
Medical records can span hundreds or even thousands of reports, 
    authored by various healthcare professionals over the years, 
    each addressing different aspects of a patient journey. 
While this ``longitudinal'' nature highlights the richness of these records, 
    it also presents substantial challenges for AI. 
AI must perform a task we call \textbf{longitudinal reasoning}, 
    which involves logical, causal, spatial (within the patient's body), and temporal reasoning. 
Additionally, it requires reasoning about functionality and behavior, such as those of organs, medications, and tumors, across extensive texts in numerous documents.


\subsection{Motivation for Longitudinal Reasoning}

Figure \ref{fig:longitudinal} illustrates the complexity of longitudinal reasoning in a patient record. 
The example includes three independent reports for a single patient: 
    the first by an oncologist discussing breast cancer metastasis to an ovary, 
    the second by a surgeon who removed the patient's ovaries and uterus, 
    and the third by a gynecologist describing a \textit{current} pregnancy
        at the time of writing.

Although these clinical events are documented independently, 
    several types of reasoning are required to connect and resolve them: 
        \textbf{spatial} reasoning to link the events involving the same organs, 
        \textbf{causal} reasoning to update the ``state of the world'' after the surgery 
            (removing the mentioned organs from the patient body),
        and \textbf{logical} and \textbf{temporal} reasoning to detect a temporal paradox, since pregnancy is impossible without the ovaries and uterus (reasoning about \textit{functionality}). 
Longitudinal reasoning must resolve this conflict 
    by reordering the events to create a \textit{coherent} journey, 
        such as placing the surgery after the breast cancer and pregnancy, 
    or by distrusting one or more of these beliefs 
        to maintain \textit{consistent} journey. 
    This requires examining other clues in the patient record, which might introduce additional logical and temporal conflicts.

Given a query asking for breast cancer patients who later had a pregnancy, 
    this patient would only be retrieved if the longitudinal reasoning discarded the surgery; any other resolution scenario would render the patient ineligible. 
It is important to note that this is a simplified example, 
    and actual patient records often contain thousands of documented events, 
    making reasoning and theorem proving highly complex and computationally intractable.

\subsection{The Case for Efficient Large-Scale Reasoning}
The core message of this work is that 
    cohort retrieval systems must achieve high retrieval quality,
        leveraging recent advances in AI, particularly in LLMs.
%
%
However, these improvements must not compromise efficiency at scale. 
Low cost per query and real-time inference are essential for the practical adoption of such systems in real-world scenarios that scale to millions of patients.


\subsubsection{The need for scalable solutions exploiting the recent advances in AI}
Many studies have been published on cohort retrieval 
    \citep{egar_diagrammatic_1992, fletcher_improving_2012, shivade_review_2014, hiob_health_2017, idnay_systematic_2021, yang_identification_2022, gao_scoping_2022, he_survey_2023, yang_machine_2023}; 
    some research work focuses on translating clinical trial eligibility criteria into machine-executable queries 
    \citep{egar_diagrammatic_1992, tu_methodology_1993, carlson_computer-based_1995, kang_initial_2015, wilcox_research_2015, yuan_criteria2query_2019, wong_scaling_2023}
    while other studies aim to identify a representative cohort from a broader population to minimize bias and improve fairness while optimizing patient data utilization 
    \citep{stubbs_cohort_2019, segura-bedmar_cohort_2019, dai_cohort_2020, birnbaum_model-assisted_2020, bairaktari_fair_2022, henzinger_leximax_2022, farrand_machine-learning_2023, chang_towards_2024}. 
In recent years, numerous works 
    \citep{chamberlin_evaluation_2020, liu_implementation_2020, liu_create_2020, soni_patient_2020} 
    have applied techniques from multidisciplinary research fields, including information retrieval, natural language processing, and machine learning to build cohort retrieval systems. 
However, achieving high-quality retrieval results in such systems while maintaining efficient execution
        on a large-scale patient database with longitudinal EMRs have yet to be fully explored. 
%

%
\subsubsection{Document retrieval is unsuitable for longitudinal data}
Unlike standard document retrieval tasks that focus on a single document 
        \citep{manning2009introduction, zhu_large_2023}, 
    cohort retrieval requires integrating data from multiple documents within a patient's EMR, 
        which may contain contradictory or evolving information 
            \citep{stubbs_cohort_2019, sarker_defining_2021}. 
This complexity necessitates retrieval systems 
    capable of quickly identifying and aggregating relevant information 
        from multiple sources into a coherent final answer. 
Most prior work focuses on documents related to a single patient visit 
    rather than the patient’s complete medical timeline, or ``longitudinal'' data, 
    which more accurately reflects the breadth of information available in a health system. 
In this study, we evaluate retrieval on full EMRs encompassing longitudinal patient journeys.
%
%

\subsection{{Approaches to Automatic Cohort Retrieval}}


To address the aforementioned challenges, we introduce a novel task 
    called \textbf{A}utomatic \textbf{C}ohort \textbf{R}etrieval (\textbf{ACR}),
    that extends clinical trial matching and cohort selection 
        to large-scale, longitudinal data. 
%
%


Large Language Models (LLMs) 
    demonstrate promising capabilities in preliminary healthcare tasks \citep{singhal2022large}. 
However, their effectiveness in more complex tasks requiring efficient large-scale execution 
    remains unclear when compared to neuro-symbolic solutions 
        that combine task-optimized machine learning models with symbolic reasoning and expert knowledge. 
Additionally, the safety and reliability of LLMs continue to be subjects of ongoing debate. 
A comprehensive understanding of the differences between LLM-based solutions and domain-specific hybrid systems 
    is crucial for the development of effective cohort retrieval systems.
%

\subsubsection{LLM approaches to ACR}
A practical approach involves using an LLM to embed and store patient documents or text passages in a vector database offline. 
Upon receiving a query, this method employs dense retrieval by embedding the query and conducting a similarity search for relevant documents. 
The cohort is then constructed from the patients associated with these retrieved documents.
We refer to this approach as `retriever-only' \citep{soni_patient_2020}.
Optionally, this retriever can be enhanced with a `reader,' 
    an LLM that further analyzes the EMRs of retrieved patients to verify their inclusion in the desired cohort. 
This approach is called `retrieve-then-read' \citep{yuan_criteria2query_2019}.
Both approaches struggle with context window size when handling longitudinal EMRs, 
    especially when attempting to maintain feasibility when querying extensive patient datasets.

%
\begin{enumerate}
\item \textbf{Context window limitation for longitudinal records.}
Some queries may require longitudinal reasoning across documents 
    in a patient's EMR that could span hundreds of reports. 
The retriever will struggle in capturing the full history of a patient in a vector. 
Similarly, for the reader to comprehend a patient’s medical journey, a large context window is necessary. 
While significant advances have been made in extending the context window \citep{su2021rope, peng2023YaRN},
    it comes at significant additional costs per patient per query.

    %
%
\item \textbf{Infeasible ACR with iterative LLM runs.}
Reading comprehension, a time-consuming process, 
    is better performed offline during corpus preprocessing rather than online at query time. 
While incorporating a reader might enhance ACR systems' quality, 
    it compromises feasibility since the iterative, per-patient prediction is performed online for every query. 
In this paper, we investigate the trade-off between efficiency and effectiveness in ACR systems.
While the retriever-only assesses the upper bound for speed, 
    the retrieve-then-read approach 
        serves as a theoretical benchmark for the highest achievable quality when cost is disregarded.
\end{enumerate}

To gain insights into the performance of various LLMs on this task, 
    we explored multiple text embeddings, 
        including OpenAI's \texttt{ada} and \texttt{text-3-large}, 
            and several embeddings fine-tuned on medical tasks (QA and SNLI).
    Since the reader is much more expensive with time and cost, we experimented with OpenAI's GPT-4 only.

\subsubsection{Neuro-symbolic approach to ACR}

In addition to the LLM-only approaches, 
    we incorporated a commercial product named Hypercube \citep{shekhar2023coupling} into our evaluation. 
This platform features a conversational AI engine tailored for healthcare professionals, 
    enabling users to query and analyze medical records 
        to generate insights for clinical research questions. 
Notably, Hypercube utilizes a neuro-symbolic architecture 
    that combines LLMs with medical ontologies and symbolic reasoning for cohort retrieval.
The inclusion of Hypercube in this evaluation 
    serves to contrast LLM-only approaches with a domain-specific solution, 
    showcasing the potential value of integrating expert knowledge into ACR systems.

While GPT-4, ada, and Hypercube are all closed-source commercial products, 
    this work does not delve into their internal mechanisms. 
Instead, we focus on the task at hand and conduct a detailed analysis of their performance, 
    providing evidence of the challenges these systems face 
        in achieving effective and efficient longitudinal and large-scale reasoning.

\subsection{Contributions}

The contributions of this work include the following:
\begin{enumerate}
    \item We define a new benchmark task we call \textbf{Automatic Cohort Retrieval (ACR)}
            emphasizing the need for systems that provide quality results at the scale of real-world datasets of millions of patients with longitudinal records. 
    \item \textbf{ACR query dataset} 
             containing \noquery{} queries from practical use cases in oncology 
                that we make publicly available.
    \item \textbf{An EMR dataset} of \nopatient{} patients across four sites 
                    encompassing \nodocument{} medical records. 
            Additionally, we introduce an \textbf{efficient AI-powered approach} to create large, human-labeled \textbf{query-patient pair datasets} to evaluate ACR. 
                We also make this gold dataset available for research.
    \item \textbf{An evaluation framework and interpretable metrics} 
        to measure the quality as well as the hallucination tendencies and set-theoretic inconsistencies of ACR systems.
    \item \textbf{An extensive evaluation and analysis} on three ACR baselines, two of which employ LLM-only techniques and one utilizes a neuro-symbolic approach.
        Additionally, we propose various techniques of stratifying queries and patients 
            for deeper examination of the weaknesses and opportunities of these ACR system.
    
\end{enumerate}

Our paper is structured as follows. 
    In Section~\ref{sec:related-work}, we review prior research related to this topic.
    In Section~\ref{sec:acr-benchmark}, we describe the ACR task and its evaluation framework.
        Additionally, we introduce a generic ACR system architecture and present three baseline approaches,
        and we describe our data annotation framework. 
    In Section~\ref{sec:experimental-results}, we evaluate the baselines and provide a detailed discussion and manual analysis.

\section{Related Work}
\label{sec:related-work}

The complexity of cohort retrieval escalates with the growing number of queries, patients, and documents per patient. 
Existing approaches to this task utilize wide range of technologies including  information extraction, LLMs, and information retrieval (IR).


The most common paradigm involves entity and relation extraction from EMRs offline
    \citep{weng_elixr_2011, kang_eliie_2017, nye_understanding_2021}, 
    then applying structured queries \citep{yuan_criteria2query_2019, fang_combining_2022}. 
Trial eligibility criteria are translated into machine executable queries 
    such as SQL or logical forms \citep{tu_methodology_1993, carlson_computer-based_1995, wilcox_research_2015, kang_eliie_2017, yuan_criteria2query_2019, liu_attention-based_2020, fang_combining_2022, dobbins_leaf_2022}.
%
%
Structured forms for eligibility criteria are produced as intermediate results, 
    making them easier for humans to review \citep{idnay_clinical_2023}. 
These systems, however, suffer low recall and generalizability 
    due to the lexical diversity of EHRs and the complex, compositional nature of clinical queries. 
They also requires extensive annotation efforts for information extraction and query translation. 

Recent research using LLMs has demonstrated impressive in-context learning capabilities in healthcare applications \citep{lee2023benefits, lee2023ai, nori2023capabilities}. 
Some studies \citep{wong_scaling_2023,devarakonda_clinical_2023} have shown that LLMs are promising for the extraction of complex matching logic of trial-eligible criteria while mitigating the cost of human annotations. 
It is unclear, however, whether these performances are maintained in unstructured, noisy, longitudinal real-world patient data.

Another notable approach adopts information retrieval methodologies for end-to-end cohort retrieval.
In particular, several studies learn to encode both patient and query for end-to-end retrieval 
    via language embedding \citep{glicksberg_automated_2017} 
    or via supervised learning from query-patient pairs \citep{zhang_deepenroll_2020, liu_attention-based_2020}. 
Some studies leverage known eligible patients to search for patients with similar medical history \citep{miotto2015case, chakrabarti_interoperable_2017}. 
Other studies, termed cohort selection \citep{chakrabarti_interoperable_2017, segura-bedmar_cohort_2019, karystianis_rule-based_2019, vydiswaran_hybrid_2019, chamberlin_evaluation_2020, stubbs_cohort_2019, stubbs_new_2019,al-garadi_automatic_2020, dai_cohort_2020, birnbaum_model-assisted_2020, bairaktari_fair_2022,henzinger_leximax_2022,yang_identification_2022,theodorou_treement_2023, chang_cohort_2022, chang_towards_2024} 
    formulate patient-trial matching as multi-label classification tasks. 
These systems mitigate the need for structured patient data, however, 
    they still suffer the same problem of supervised learning 
        in the lack of training data and low generalization. 

The emergence of LLMs in information retrieval \citep{zhu_large_2023} 
    has enabled enormous potential for cohort retrieval. 
Recent work \citep{zhu_large_2023,jin_matching_2023,nievas_distilling_2023,kusa_effective_2023} 
    have shown promising results    
    leveraging LLMs embeddings for retrieval and LLMs’ in-context learning ability for patient-trial matching. 
LLMs, however, have some limitations that have yet to be solved 
    such as feasibility on large corpora while maintaining effective reasoning capabilities.


\section{ACR Benchmark}
\label{sec:acr-benchmark}
Recent progress in information retrieval (IR) and LLMs 
    has unlocked new possibilities for cohort retrieval,
    necessitating a benchmark that emphasizes \textit{quality at scale} for real-world applications.
%


In IR, there is a notable trend toward addressing more complex queries that span multiple documents. 
Existing datasets like HotpotQA, while demanding multi-document reasoning, have limited scope, usually focusing on only one or two paragraphs per query.
Recent developments, however, has shifted towards more complex benchmarks that require the analysis of several documents per query. 
Such benchmarks include AQuaMuse, HowSUMM, and WikiHowQA.
Cohort retrieval shares common NLP challenges with other IR tasks.
It demands fast retrieval of results while maintaining high precision and recall.
As queries become more complex, cohort retrieval necessitates complex multi-document reasoning. 
This is because there are often several criteria in the query with their answers spread across multiple documents in a patient's EHR. 
%


\subsection{Task Definition}
\label{sec:task_definition}

We propose a benchmark for Automatic Cohort Retrieval (ACR).
Formally, consider a corpus of patients $\mathcal{P} = \{(i, r_i)\}$,
    where $i$ is a patient identifier 
        paired with her corresponding medical record 
            consisting of a collection of text documents, $r_i = \{ d_{ij} \}$. 
Given a query $q$, the task of cohort retrieval 
    aims to produce a cohort $\mathcal{C} \subset \mathcal{P}$ as an answer to $q$. 
A patient $i$ is eligible for a query $q$ if $r_i$ provides evidence 
    that patient $i$ satisfies all the inclusion and exclusion criteria defined in $q$,
        denoted as $r_i \implies q$. 
We emphasize that the focus of this task is to build real-world ACR systems that can scale to millions of patients each with longitudinal records. 

\textbf{Oncology Focus.}
Medical records are dense in information and jargon, are frequently repetitive and ambiguous, and reflect dynamic conditions of disease and its treatment evolving over time. 
In oncology, disease journeys can span many years across multiple treatment episodes and disease states. 
Cohort retrieval in this context poses significant challenges for both systems and human experts.
Our benchmark begins with this most difficult case, under the assumption that systems that can perform well in this setting can be generalized to simpler cases in other therapeutic areas. 



\subsection{Evaluation Framework}
\label{sec:eval}



\subsubsection{Evaluated Phenomena}
\label{sec:eval-phenomena}

In order to appropriately assess an ACR system's effectiveness, 
    we propose an ensemble of evaluations to examine the following phenomena: 
        

\begin{enumerate}
     \item 
\textbf{Retrieval Quality.}
        As patient records can span many visits and types of documents, it is important for an ACR system to be able to accurately understand the interaction between various conditions and interventions over time. 
As the number of documents in a patient record grows, the task of tracking clinical events across documents, aggregating their evolving facts while resolving any emerging conflict, becomes more challenging. 
This longitudinal reasoning obstacle hinders the system's ability, 
    at query time, 
    to correctly identify the necessary information crucial to retrieve or reject a patient,
        resulting in various false positives and negatives.

     \item 
 \textbf{Hallucination.}
The term \textit{hallucination} has been historically used by the machine translation community and was adopted later by LLM literature to refer to the production of unfounded or non-factual responses. 
In this work we advocate for extending the use of \textit{hallucination} to the domain of information retrieval (IR),
    defined as the phenomenon when an IR system retrieves false positives -- results that are factually incorrect in relevance to the query.
Hallucinations become most detrimental in searches for patients with rare combinations of clinical events,
    where queries might yield no or minimal results. 
    Consequently, when the patient population size escalates to millions, 
        manual verification of cohort membership becomes impractical, even with relatively low hallucination rates. 

     \item 
\textbf{Set-Theoretic Consistency.}
We define two types of set-theoretic consistency checks for an ACR system.
First, we introduce \textbf{Paraphrased queries}, 
    which are queries that convey the same meaning but express it through different synonyms or paraphrasing. 
The cohorts returned for such queries must be equivalent sets. 
For example, the following two queries should result in an identical cohort of patients:
    ``non-invasive breast cancer who underwent a lumpectomy''
    and 
    ``patients with malignant breast neoplasm, in situ, treated surgically with a segmental mastectomy''
The second type of consistency check involves \textbf{Complex queries} that combine different criteria 
    using set operations such as intersection, union, and difference 
    through terms like ``and,'' ``or,'' and ``except,'' respectively. 
For example, the query ``female lung cancer patients'' 
    represents an intersection between two otherwise unrelated queries: 
    ``female patients'' and ``lung cancer patients.'' 
To ensure the ACR system is \textbf{\textit{consistent}}, 
    the result of this intersection must not contain any patients outside of the two sub-queries.
Additionally, we define \textbf{Subtype queries} that relate a parent query to a child query. 
For example, consider the following three queries:
    ``cancer patients'', ``liquid tumor patients'', and ``leukemia patients''.
Since leukemia is a type of liquid tumor which is a type of cancer, 
    each of these queries is a subtype of the previous one. 
A subtype query must have a cohort that is a strict subset of its parent query. 
This set-theoretic consistency evaluation is \textit{inherently unsupervised}, eliminating the need for ground truth labels which could scale the set of queries to millions.

 \end{enumerate}


\subsubsection{Query Categorization} \hfill
\label{sec:data_cats}

\begin{enumerate}

\item \textbf{Data-driven categorization.}
Various classes of queries arise from a combination of the queries and dataset considered. 
More explicitly, accuracy metrics for queries that have only a few, or no, patients 
    carry a different weight than for those that return many. 
Thus, aggregating these scores together can result in misleading characterization of a system's abilities. 
We organize queries into 4 categories determined by their ground truth cohort size, $n$:
 \begin{enumerate}
     \item 
 \textbf{Broad} spanning queries where $n \geq \alpha$, 
     \item 
 \textbf{Narrow} spanning queries returning  $ n \in [\beta, \alpha)$, patients
     \item 
 \textbf{Sparse} queries with $n \in [1, \beta)$ patients retrieved, and 
     \item 
 \textbf{Zero-Result} queries returning no patients at all.
 \end{enumerate}
Here, $\alpha$ and $\beta$ are determined by the number of patient in the corpus
($\alpha = 50$ and $\beta = 10$ in our 1.4K-patient corpus).

\item \textbf{Expert-driven categorization.}
%
Estimates of clinical complexity were produced as an alternative subjective measure of query complexity. 
Clinical experts assessed and categorized queries by: 
\begin{enumerate}
    \item the number and kind of clinical observations required to identify the cohort,
    \item the degree of biomedical specialization involved in interpreting the query properly, and 
    \item overall length of the query
\end{enumerate}
%
For example, \textit{``patients with lung cancer"} is considered a \textbf{Base} case level of complexity as it involves one condition widely known to non-specialists. 
\textbf{Low} complexity queries like, \textit{``patients with advanced NSCLC"}, implicitly involve multiple conditions; however, \textit{"advanced NSCLC"} is a frequent concept in medical text, and requires a low degree of medical specialization to understand. 
Next, \textit{``patients with HR+/Her2- breast cancer and a mutation in PIK3CA"} involves several criteria often spread across a patient record and requires a moderate degree of biomedical specialization to interpret (\textbf{Medium} queries). 
Lastly, queries with a significant length, number of conditions, nested logic, and specialization needed to understand the query are \textbf{Hard} queries; 
    e.g., \textit{``adults with locally advanced or metastatic solid tumor or multiple myeloma, no primary brain tumors or leptomeningeal metastases"}. 
%
These measures are provisional and we welcome further refinement of rigorous formulation of clinical query complexity.

\end{enumerate}

\subsubsection{Metrics}

To evaluate ACR on the query categories above, 
    we follow established prior work, leveraging \textit{macro} Precision, Recall, and F1-score (P/R/F1) for \textit{Broad} and \textit{Narrow} queries \cite{wong_scaling_2023}. 
However, due to the sensitivity of these metrics for \textit{Sparse} and \textit{Zero-Results} queries (division by zero), \textit{micro} P/R/F1 are reported for Sparse queries, 
    while hallucination analysis is performed on the Zero-Result queries.
\subsection{Generic ACR System Architecture}

\begin{figure*}[t!]
    \centering
    \includegraphics[width=\textwidth]{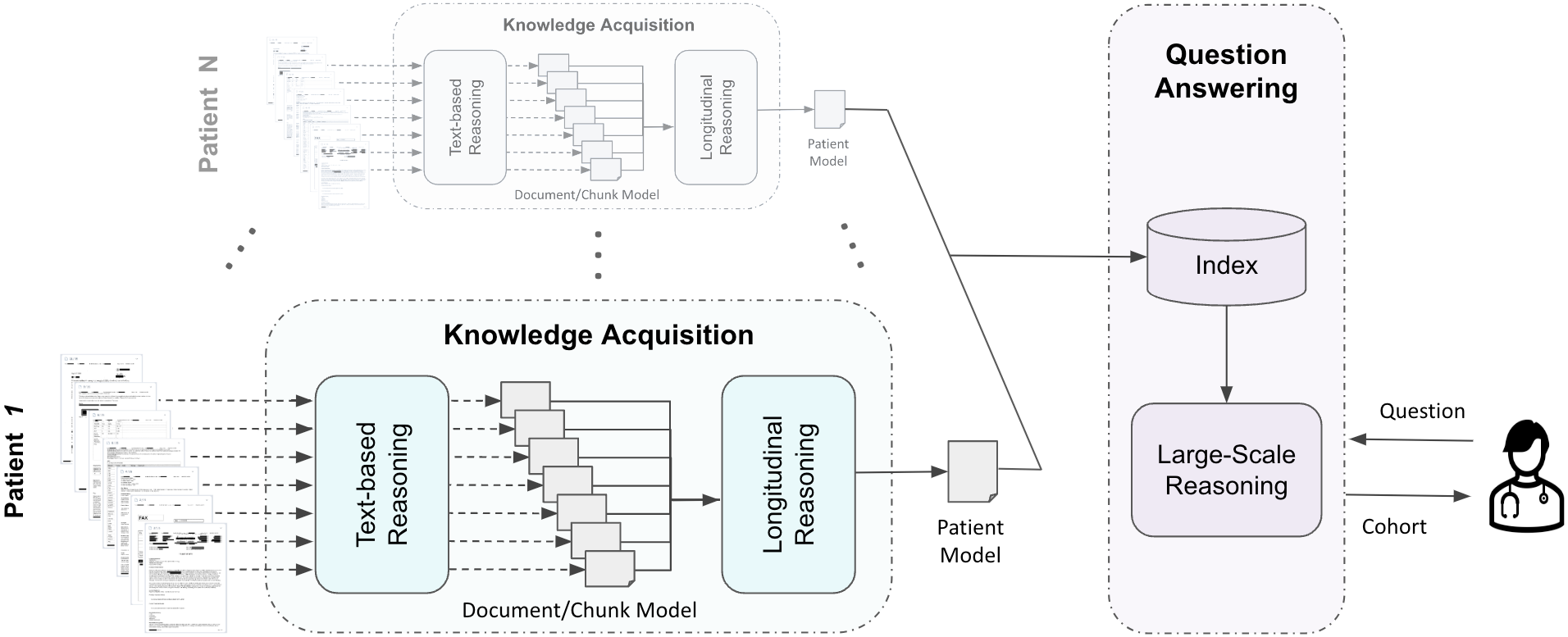}
    \caption{Architecture of a generic ACR system:
        Given a query, \textit{large-scale reasoning} is conducted 
            over numerous patients with longitudinal EMRs. 
    This involves \textit{text-based reasoning} on document or chunks, followed by \textit{longitudinal reasoning} over time.}
    \label{fig:overview}
\end{figure*}

\subsubsection{Indexing and Interrogation in Information Retrieval.}
An IR system integrates online and offline processes that play crucial roles in data organization and retrieval efficiency. 
\begin{enumerate}
    \item 
\textbf{Indexing}, performed \textit{offline}, involves pre-processing tasks like parsing, tokenization, stemming, and the construction of inverted indexes to create a structured representation of data. 
These processes are performed in advance to optimize the system for swift information retrieval. 
\item 
\textbf{Interrogation}, performed \textit{online}, is executed in real-time and involves the dynamic querying of the system based on a user query. 
This includes activities such as query parsing, expansion, matching against the index, ranking based on relevance, and the final rendering of information. 
\end{enumerate}
Together, these processes ensure that the system incurs the cost of processing a document only once during the offline phase, 
    while retaining the ability to retrieve it for an unlimited number of online queries without requiring reprocessing.

\subsubsection{ACR Architecture as an IR System}

Similar to IR systems, an ACR system integrates two primary processes: 
\textit{Knowledge Acquisition} (KA) and \textit{Question Answering} (QA), 
as depicted in Figure~\ref{fig:overview}.

\noindent\textbf{(A) \hspace{0.01cm} Knowledge Acquisition (KA).} 
    Analogous to indexing, KA is tasked with transforming an extensive patient medical record, containing hundreds of reports written over several years, into a Patient Model, representing the patient's clinical journey. This process unfolds in two phases:
    



 \begin{enumerate}
 \item
    
    \textbf{Text-based Reasoning:} This phase employs Natural Language Processing (NLP) and Understanding (NLU) to convert medical text into a document or chunk model. 
    It faces challenges due to the limited context window NLP systems can process at once and the heterogeneous nature of patient records, 
        which comprise documents authored by various clinicians at different times, 
        each with temporal references specific to their writing dates. 
    Consequently, this phase processes an EMR one document or chunk at a time, relying on subsequent steps for fact consolidation and temporal resolution.

    \item
    \textbf{Longitudinal Reasoning} is the capacity to comprehend and retain knowledge from an extended narrative that develops over time. 
    It entails tracking the progression of events and their facts, resolving any contradictions, and piecing together evolving but compatible elements into a consistent and comprehensive journey while maintaining coherence and explainability. 
    This phase consolidates the document-level models into an all-encompassing patient model. 

 \end{enumerate}

\noindent \textbf{(B) \hspace{0.01cm} Question Answering (QA).}
Analogous to interrogation, QA addresses the retrieval of patient cohorts based on specific inclusion and exclusion criteria. 
Given the potential for databases to contain millions of patient records in any practical healthcare application, efficient retrieval is facilitated by:
\begin{enumerate}
    \item A \textbf{storage-efficient index} that organizes patient models for quick access and retrieval.

    \item A \textbf{large-scale reasoning} mechanism over the index capable of efficient and accurate cohort retrieval. 
    
\end{enumerate}

\subsection{Baseline Approaches}
\label{sec:approach}

In this section, we evaluate three approaches to ACR varying in their underlying technologies 
    and we discuss their efficiency and effectiveness. 

\subsubsection{Retriever-only.}
\label{sec:baselines-retriever}

This approach employs dense retrieval on segmented text passages. 
The offline knowledge acquisition process 
    segments patient records into passages of text 
    and embeds them using an LLM, 
    storing them in a vector database. 
During the online QA process, user queries are similarly embedded and dense retrieval is used to find relevant passages in the indexed corpus. 
Given that the ACR task focuses on retrieving a cohort of patients rather than individual text passages, 
    these passages are grouped by patient ID to form the resulting patient cohort.

This approach is computationally efficient as the KA process is performed offline, 
    only once per passage, while the online process performs an efficient similarity search. 
However, it lacks longitudinal reasoning, 
    which is crucial for queries that require piecing together clues spread across multiple passages 
    especially when the chronological order of clinical events is important to answering the query. 
Additionally, while this dense retrieval method is efficient for large-scale reasoning, 
    it may not be sufficiently effective. 
    It relies on the assumption that the independent embeddings of a passage and a query contain all necessary information for effective reasoning. 
    This assumption may fail for queries with vague terms, negation, or diverse criteria, and for passages with hints, vagueness, or uncertainty. 
    Furthermore, Euclidean embeddings struggle with representing hierarchies, leading to vague queries being embedded too far from relevant passages, potentially missing crucial information, or too close, retrieving similar but irrelevant passages.
    
\subsubsection{Retrieve-then-read.} 
\label{sec:baselines-retrieve-then-read}

To address the challenges of the Retriever-only approach, we consider the Retrieve-then-read baseline.
This approach utilizes the same retriever;
however, during online QA, it follows the dense retrieval step with a ``reader,''
    which is an LLM that is prompted by the user query along with the top $k$ retrieved passages per patient 
        and is asked to give a yes/no answer regarding whether the patient belongs to the query's cohort. 

While this approach cannot improve the Recall of the retriever, it may be able to enhance its Precision 
    since the reader is jointly examining the query and the passages,
        performing longitudinal reasoning online. 
While this allows for more effective reasoning, 
    it is computationally infeasible in practice
    since it requires prompting an LLM-based reader for each retrieved patient for every query. 
    In any realistic EMR setting comprised of millions of patients, the cost of a single query will be financially prohibitive.
Even though this solution is infeasible, we still consider it to measure the upper limit on LLM-based reasoning, referring to it as an \textit{oracle}. 


\subsubsection{Neuro-symbolic approach (Hypercube).}
\label{sec:baselines-hypercube}

For this approach, we evaluate a commercial product named Hypercube that performs cohort retrieval and analysis.
Hypercube pairs LLMs with symbolic reasoning \citep{shekhar2023coupling}
    in the following components depicted in Figure~\ref{fig:overview}: 
 \begin{enumerate}
     \item
    \textbf{Text-based reasoning:} Each medical report is fed to a lossless clinical NLU process that interprets both the explicit text content and its tacit implications at a document level, and faithfully transforms it into a proprietary Knowledge Representation (KR).

     \item 
    \textbf{Longitudinal reasoning:} Hypercube utilizes a Dynamic Symbolic Memory as it examines all KR artifacts extracted from each report and cross-references them against the evolving journey in the dynamic memory. 
    When new evidence aligns with current knowledge, the engine consolidates events and refines their facts, boosting their confidence scores. 
    Conversely, if evidence conflicts, the engine scrutinizes all contradictory beliefs within the journey, leveraging multiple symbolic AI and ML resources to ascertain the most plausible beliefs, updating or amending the memory as necessary. 
    The result is a comprehensive, consistent, and coherent symbolic journey where each belief is both justified and traced to its originating text.

    \item 
    \textbf{Large-scale reasoning:} A proprietary event-rooted knowledge base is developed to allow efficient and effective reasoning across millions of patients to answer clinical queries. 
    %
    Our study limits its evaluation to cohort retrieval, which is only one type of query supported by Hypercube.
    Additionally, Hypercube ranks the results in the retrieved cohort and provides explanations for the inclusion of each patient; both tasks are outside the scope of this paper. 

    \item 
    \textbf{Interface:} Hypercube features a \textbf{\textit{symbolic query language}}, named \textit{Eloquent}, allowing users to effectively express any query supported by the KR. 
    The execution of Eloquent over the knowledge base for large-scale reasoning is mathematically provable. 
    Alternatively, an LLM-based language user interface (LUI) is provided to translate natural language queries into Eloquent. 
    Furthermore, users can use SQL to query Hypercube; internally, SQL is translated to Eloquent to allow for large-scale reasoning.

 \end{enumerate}

\subsection{Data Annotation Framework}
\label{sec:data_anno}


\subsubsection{Query Dataset.}
The ACR task uses a dataset of \noquery{} complex queries written by medical experts on \nousecase{} precision oncology use cases \citep{schwartzberg_precision_2018} listed in Appendix~\ref{apd:checklist}. 
%
%
%
We are making this query dataset publicly available and we are attaching it in Appendix~\ref{apd:question-bank}.


\subsubsection{Labeled Patient Dataset}

Any useful \textit{query-patient pair} gold dataset should contain at least hundreds of thousands of pairs to cover various phenomena in queries as well as in patient records. 
The patient data we aggregated consists of de-identified unstructured medical records pooled from \nosite{} large academic or community oncology practices for a representative sample of \nopatient{} patients containing \nodocument{} individual reports or documents. 
These sites were chosen to represent diversity in patient population and reporting and documentation types so that it would not be possible to simply memorize a scheme, which could be used for trivial extraction.
%
%
Patient records in this set 
    have \avgnumdoc{} documents on average, 
        rendering it impractical and prone to error to manually annotate all queries against all patients 
    in a gold matrix with a \nopatient{} $\times$ \noquery{} $=$ \nocells{} cells. 

In this work we present an efficient approach to create such large gold matrices.
First, clinical experts perform a task called ``clinical abstraction,'' 
    where they read every patient record and extract clinical facts into a data model that covers all variables mentioned in our query dataset.
Second, we use the deterministic large-scale algorithm used in Hypercube 
    to create the gold matrix containing all query-patient pairs and their labels. 
    For that purpose, we load the gold patient abstraction dataset into Hypercube's knowledge base, 
        and, then, we run each query through Hypercube's deterministic reasoning engine 
            to obtain the associated cohorts as ground truth. 
%

    \begin{table}[ht]
\setlength{\tabcolsep}{1pt}
\caption{Evaluation of cohort retrieval systems on precision oncology queries categorized by the size of the gold cohort.}
\label{tab:retrieval_results}
\centering
    \begin{tabular}{lP{2cm}P{2cm}P{2cm}cP{2cm}P{2cm}P{2cm}}
    \toprule
    \centering
    \multirow{2}{*}{Models} & \multicolumn{3}{c}{Cohort Retrieval} & & \multicolumn{3}{c}{Oracle Top-$k$} \\ 
    \cline{2-4}
    \cline{6-8}
    & Precision & Recall & F1 & \hspace{0.35cm} & Precision & Recall & F1 \\
    \midrule
    \multicolumn{8}{c}{\textbf{Broad Queries}} \\
    \multicolumn{8}{c}{{\footnotesize Macro Average (18 queries)}} \\
    \midrule
    ada         & 61.59         & 56.71         & 52.07         & & 62.34           & 59.99         & 61.14         \\
    ada+GPT4    & \textbf{79.38}& 47.09         & 52.68         & & \textbf{81.08}  & 48.80         & 58.37         \\
    Hypercube   & 76.19         & \textbf{87.43}& \textbf{79.40}& & 78.84           & \textbf{74.95}& \textbf{77.13}\\
    \midrule
    \multicolumn{8}{c}{\textbf{Narrow Queries}} \\
    \multicolumn{8}{c}{{\footnotesize Macro Average (13 queries)}} \\
    \midrule
    ada         & 10.68         & \textbf{68.29}& 16.84         & & 32.21           & 31.60         & 31.90         \\
    ada+GPT4    & 38.34         & 50.68         & 41.23         & & 54.30           & 28.94         & 35.50     \\
    Hypercube   & \textbf{56.10}& 54.83         & \textbf{52.02}& & \textbf{57.84}  & \textbf{47.69}& \textbf{51.24}\\
    \midrule
    \multicolumn{8}{c}{\textbf{Sparse Queries}} \\
    \multicolumn{8}{c}{{\footnotesize Micro Average (41 queries)}} \\
    \midrule
    ada         & 1.14 & \textbf{78.95} & 2.25 & & 24.77 & 23.68 & 24.22 \\
    ada+GPT4    & 17.87 & 50.00 & 26.33 & & 59.09 & 22.81 & 32.91 \\
    Hypercube   & \textbf{32.64} & 41.23 & \textbf{36.43} & & \textbf{65.62} & \textbf{36.84} & \textbf{47.19} \\
    \bottomrule
    \end{tabular}

\end{table}


To validate the mathematical correctness of Hypercube's reasoning engine 
    and ensure it is not introducing any bias in favor of Hypercube over the other baselines,
    we validated the quality of the ground truth 
        by randomly selecting a subset of 20 queries and 50 patients 
        resulting in 1,000 pairs to be annotated manually by medical experts. 
We compared the 1,000 human-labeled query-patient pair labels 
    and we observed a Cohen's Kappa coefficient of \kappascore{} 
        suggesting perfect agreement between the automatically-generated labels and expert-annotated ones. 
This additional quality control is necessary to confirm the deterministic execution of Hypercube's reasoning engine. 
%
Given the perfect agreement between these two approaches, 
    it is reasonable to assume that the Hypercube-queried knowledge base approach 
        produces accurate ground truth cohorts 
            and does not give any unfair advantage to the neuro-symbolic baseline. 

Traditional gold matrix curation 
    involves having humans read every patient record for every query 
        to make cohort inclusion decisions, 
    which can lead to errors and inconsistencies. 
Our automated approach
    ensures high-quality query-patient pairs
    since clinical abstractors process each patient once, 
        extracting clinical information and mapping it to a medical ontology, 
    which allows for multiple rounds of review and guarantees consistency across queries.

\section{Experimental Results}
\label{sec:experimental-results}
This section presents the performances 
    of the three methods discussed in \S\ref{sec:approach} on the proposed ACR benchmark. 
We implement baselines of these three approaches 
    and describe their details in \S\ref{sec:baselines}. 
We discuss the results according to the proposed evaluation framework described in \S\ref{sec:eval}.

\subsection{Experimental Setup of Baselines}
\label{sec:baselines}


\subsubsection{Retriever-only.}

We consider the following text embeddings: \href{https://openai.com/blog/new-and-improved-embedding-model}{\textbf{ada}}, \href{https://openai.com/blog/new-embedding-models-and-api-updates}{\textbf{text-embedding-3-large}}, \href{https://huggingface.co/sentence-transformers/all-mpnet-base-v2}{\textbf{SBert}} \citep{song2020mpnet}, \href{https://huggingface.co/TimKond/S-PubMedBert-MedQuAD}{\textbf{PubMed$_{\text{QA}}$}},  \href{https://huggingface.co/pritamdeka/PubMedBERT-mnli-snli-scinli-scitail-mednli-stsb}{\textbf{PubMed$_{\text{SNLI}}$}} \citep{deka2022evidence}, \href{https://huggingface.co/TimKond/S-BioLinkBert-MedQuAD}{\textbf{BioMed$_{\text{QA}}$}} and \href{https://huggingface.co/pritamdeka/BioBERT-mnli-snli-scinli-scitail-mednli-stsb}{\textbf{BioMed$_{\text{SNLI}}$}} \citep{deka2022evidence}. 
To accommodate for the context window size limit, all patient documents are split into chunks of $1K$ tokens with a $10\%$ overlap for all tested retrievers. 
The resulting index has $472,861$ embedded chunks in total. 


\subsubsection{Retrieve-then-read.} 
We evaluated different embeddings and found \textbf{ada} to outperform all others. 
We report results on \textbf{ada} as the retriever 
    along with \textbf{GPT4} \citep{gpt4_tech_report} as the reader, 
    and refer to this baseline as \textbf{ada+GPT4}. 
We empirically pick top-$k$ retrieved chunks per query to be 1,000 chunks ($0.21\%$ of our index size). 
We use GPT4 with a 128K context window and \href{https://platform.openai.com/docs/api-reference/chat/create}{default parameters}. 
Unlike the Retriever-only approach, this approach is unscalable when the number of patients grows beyond a few thousands, making it infeasible for real-life applications.
Nevertheless, evaluating this approach is essential to assess the upper limits of LLM longitudinal reasoning.


\subsubsection{Neuro-symbolic.} 
We run Hypercube with its default configurations. 
Hypercube runs once on the raw unstructured text to build a knowledge base offline from every patient EHR \citep{shekhar2023coupling}, 
    performing text-based and longitudinal reasoning. 
At query time, Hypercube reasons over the knowledge base 
    and retrieves eligible patients.

\subsection{Retrieval Quality}
\label{sec:ret_results}

    \begin{figure*}[t]
    \vspace{-0.3em}
    \centering
    \includegraphics[width=1\textwidth]{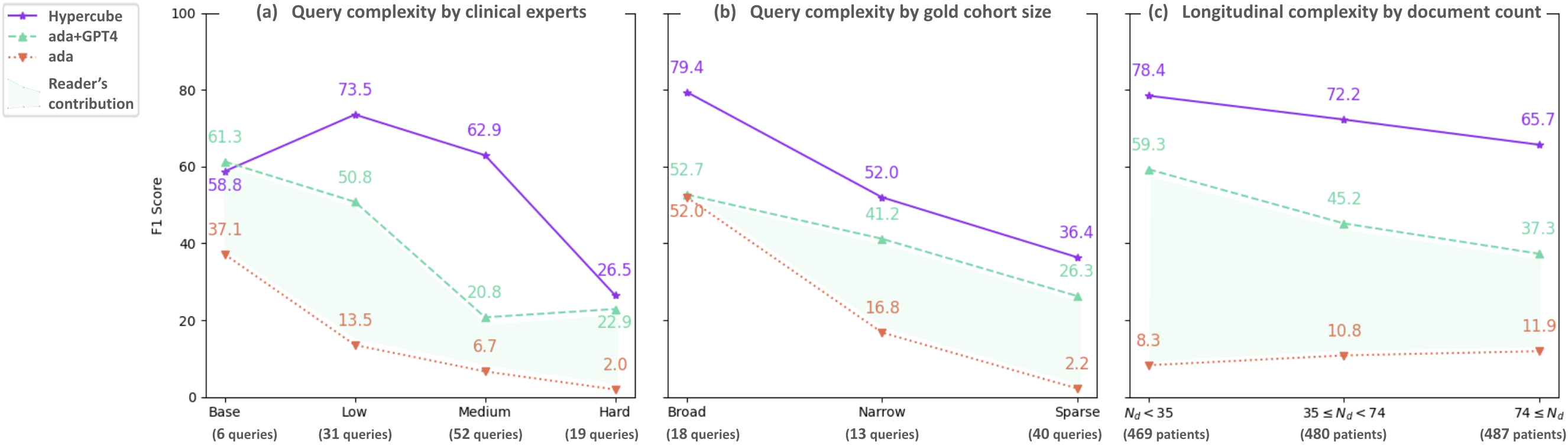}
    \caption{
        F1-Score stratified by complexity determined by (a) experts, (b) cohort size, and (c) document count per patient ($N_d$).}
    \label{fig:combined-results}
\end{figure*}


\subsubsection{Retrieval Evaluation}
As discussed in the data-driven categorization \S~\ref{sec:data_cats}, 
    we classify the queries into 4 categories.
In our dataset of ~1.4K patients, we have $\alpha = 50$ and $\beta = 10$ resulting in
    \textbf{Broad} spanning queries (18 queries) where $n \geq 50$, 
    \textbf{Narrow} spanning queries (13 queries), returning  $ n \in [10, 50)$ patients, 
    \textbf{Sparse} queries (40 queries), with $n \in [1, 9]$ patients retrieved, and 
    \textbf{Zero-Result} queries (42 queries), which return no patients at all.
Table~\ref{tab:retrieval_results} shows the performance of 
    the best retriever-only baseline (ada), 
    the retrieve-then-read baseline (ada-GPT4) 
    and the neuro-symbolic baseline (Hypercube) 
        in terms of Precision, Recall, and F1-Score (P/R/F1)\footnote{Full results with all baselines are presented in Appendix \ref{apd:results}.}. 
The table is divided into 3 sections representing the first 3 query categories\footnote{ 
    The Zero-Result category cannot be evaluated in terms of P/R/F1 due to division by zero.
    We evaluate Broad and Narrow queries using P/R/F1 \textit{macro} averages 
        and Sparse queries using \textit{micro} averages.
    This is due to systems often retrieving no true positives to queries with one or few gold answers 
        resulting in zeros or divisions by zeros for P/R/F1.
    }.
Figure~\ref{fig:combined-results}, part (b), displays F1-Score against the three categories.
Across all three query categories, 
the neuro-symbolic approach (Hypercube) consistently outperforms the LLM-only baselines on F1-scores ($10.1\%$ to $26.72\%$ gap). 
This indicates that domain-specific knowledge as well as performing journey consolidation offline 
    has a significant effect on ACR quality.
The retriever-only baseline tends to over-retrieve patients 
    which often results in a higher recall than ada+GPT4 
    but at the cost of a much lower precision. 
On the contrary, the retrieve-then-read baseline drastically improves precision but at a fair sacrifice of recall, resulting in overall much better F1-score. 
The green area in Figure~\ref{fig:combined-results}, part (b), 
    depicts the F1 contributions of the reader over the retriever.
This consistent behavior shows that the ``reader'' 
    is performing effective text-based and longitudinal reasoning online 
    when given the query along with the retrieved text passages,
    effectively second-guessing the retriever as well as attending across passages if needed to answer a query.
%

\subsubsection{Ranking Analysis}
While this paper focuses on retrieval quality, 
    we also aim to assess the quality and potential improvement gap of \textit{result ranking}.
We consider only the top-$k$ retrieved patients for measuring P/R/F1, 
    where $k$ is the size of the gold cohorts,
    assuming the baselines were given the size of the gold cohort along with the query.
Table~\ref{tab:retrieval_results} presents the results under the \textit{Oracle Top-$k$} part. 

Both LLM-only approaches show significant improvements when given the number of expected results; 
    however, it is still not enough to match the neuro-symbolic system.
%
Overall, Hypercube loses more in Recall than what it grains in Precision for macro averages
    indicating that its ranking capabilities can be significantly enhanced.



\subsubsection{The effects of longitudinal data.} 
In this section, we study the effect of longitudinality of medical records on the degradation of model performance.
We divide patients into three equal groups: 
    the bottom third ($N_d < 35$), 
    the middle third ($35 \leq N_d < 74$).
    and the top third ($N_d \geq 74$).
%
Figure \ref{fig:combined-results}, part (c), illustrates the changes of models' F1-scores as patient records grow in the number of documents per patient (denoted as $N_d$). 
Hypercube and ada+GPT4 show clear quality degradation as patient records get longer\footnote{
    One exception to these results is the retriever-only baseline, ada, 
        which seems to improve with longitudinality.
    An exception to this is the retriever-only model, ada, 
        which seems to improve with longitudinality. 
    This improvement occurs because ada starts with a low F1 of 8.3\%, 
        and the repetitions in longitudinal records provide multiple opportunities for ada to retrieve a patient
            making it easier.
};
however, ada+GPT4 degrades more than Hypercube. 
This may be because Hypercube employs offline longitudinal reasoning to examine all patient records, 
    whereas GPT4 only examines the passages retrieved by ada online.
%
%
These results highlight the importance of building comprehensive longitudinal reasoning offline over the whole medical record. 
%


%

The challenge for offline reasoning is that the queries will not be available for a joint embedding or targeted extraction. 
Hypercube claims to have a comprehensive \textit{world model} of medicine against which its knowledge base is constructed offline from patient records regardless of length. 
While the results presented in this section show Hypercube significantly outperforms LLM-only baselines and suffers the least performance degradation, 
    they also indicate there is still a long way to go to improve ACR systems' performance on longitudinal records. 
Furthermore, these findings suggest that ACR datasets with short records 
    might be too easy and unrepresentative of the actual complexity of the ACR task in the real-world.


\subsubsection{Expert-Driven Categorization}

Figure~\ref{fig:combined-results}, part (a), displays F1-scores for the three baselines 
    across the expert-driven categorization outlined in \S\ref{sec:data_cats}. 
The results show a clear trend in performance, 
    consistent with what experts predicted for the Low, Medium, and Hard difficulty queries. 
An exception to these results is Hypercube performing 2.5\% lower than ada+GPT4 
    on the Base class (macro average over only 6 queries).
In contrast, in the Medium category (52 queries), Hypercube outperforms ada+GPT4 by 62.9-20.8=42.1\% absolute, or 3 times better relative.
%
%
%
The ada+GPT4 baseline outperforms ada
     in all 4 categories showing the value of the reader.

The alignment between expert predictions and system performances 
    serves as a valuable sanity check for system development. 
Additionally, it highlights that Hard queries are significantly more challenging than the other classes. 
Consequently, future query datasets should include a greater proportion of Hard queries. 
Moreover, further research is needed to explore methods for parsing or breaking down these complex queries into simpler components.


\subsubsection{Conclusion on Quality and Efficiency} 
As shown in all parts of Figure~\ref{fig:combined-results}, 
    the methods for query and patient stratification 
        align with our expectations for model quality. 
These detailed evaluations reveal various weaknesses in ACR systems that need to be addressed. 
They also guide dataset curators 
    to select Hard and Sparse queries more often
    and to curate patient corpora with longitudinal records.

The trade-off between effectiveness and efficiency remains a significant challenge in developing practical ACR systems.
The read-then-retrieve baseline demonstrates significant quality gains 
    but incurs a substantial increase in computational cost.
%
%
To mitigate this, we limited the API calls per patient per query to three 
    due to OpenAI's rate limits at the time of this research.
%
%
This approach costs around \$$0.10$ per retrieved patient for every query (with GPT4), 
    rendering it infeasible to scale, 
        since a single query over a million patients could cost up to \$100K. 
For Hypercube, the cost is minimal 
    since it is incurred only once during the offline pre-processing of all patients into its knowledge base. 
Online, a query takes an average of 20 milliseconds 
    on a machine with 16 CPUs and 32GB of RAM.
%




\begin{figure*}[t!]
    \vspace{-0.3em}
    \centering
    \includegraphics[width=\textwidth]{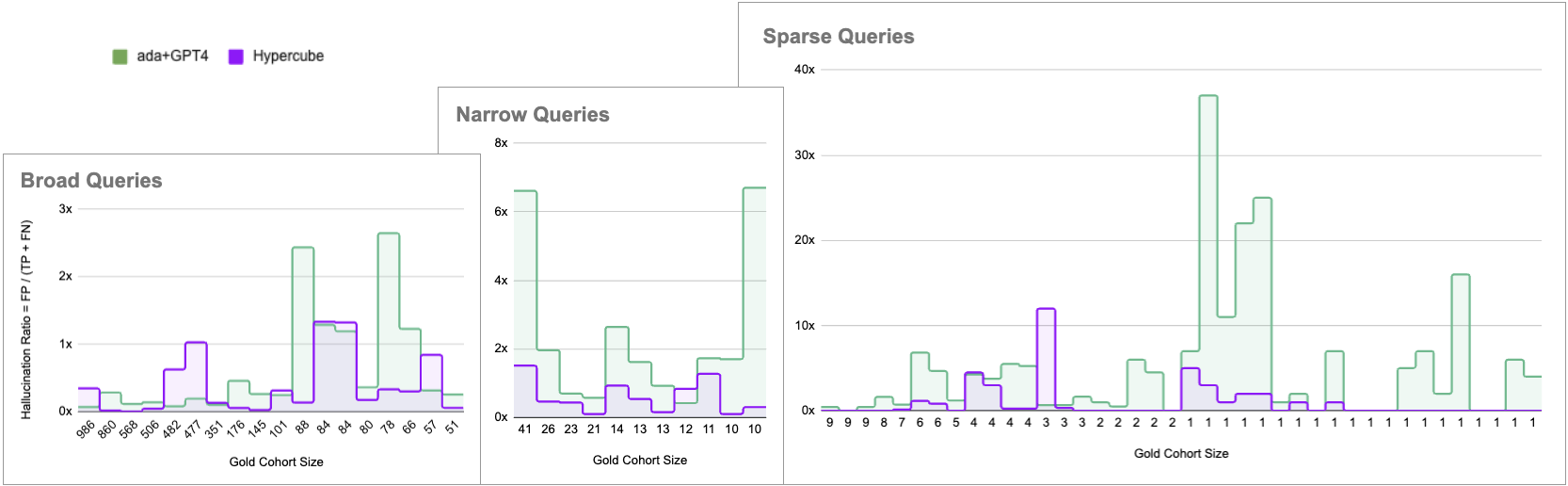}
    \caption{
    A plot of hallucination ratio against gold cohort sizes for all queries of the \textit{Broad}, \textit{Narrow}, and \textit{Sparse} types.
    It helps identify queries where models over-hallucinate, retrieving too many unqualified patients relative to actual qualified ones.
    }
    \label{fig:hallucination-ratio}
\end{figure*}

    \begin{figure}[t]
    \centering
  \includegraphics[width=.8\columnwidth]{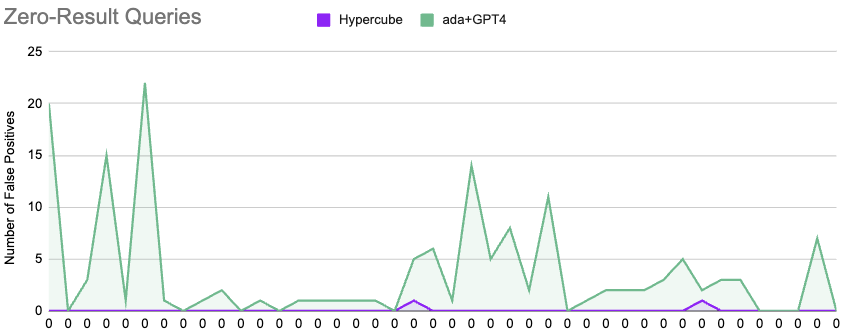}
  \vspace{-1em}
  \caption{Hallucination tendencies on every Zero-Result query (\textit{x-axis}) measured as False Positives counts (\textit{y-axis}).}
  \label{fig:hallucination}
\end{figure}


\subsection{Hallucination Tendency}

\textit{Hallucination} is 
    a common phenomenon in generative AI systems, 
        referring to instances where the model generates content 
            that is factually incorrect or entirely fabricated.
To the best of our knowledge, this phenomenon
    has not been discussed in the context of information retrieval tasks. 
In this section, we introduce a metric to gauge the hallucination tendencies of ACR systems.

\subsubsection{Hallucination Ratio.}
False positives (FPs), 
    defined as patients incorrectly included in a cohort, 
    can be thought of as hallucinations of that ACR system. 
We introduce \textbf{hallucination ratio} (HR), 
    defined as $HR = \frac{FP}{TP + FN}$, 
    which measures the ratio of FPs to all \textit{actual answers} according to gold. 
HR is a ratio, not a percentage, and can be thought of as the number of hallucinations allowed relative to the total number of actual answers.

The inspiration behind HR comes from user tolerance to hallucinations 
    based on their expectations 
    given the breadth or specificity of their query. 
For broad queries, a system needs to cast a wide net to capture relevant patients,
    and it is expected to catch some noise relative to the number of actual patients. 
Conversely, for very specific queries with few expected answers, 
    users anticipate very few hallucinations.
HR differs from precision in that it includes false negatives in its formula. 
Since FNs are actual gold answers, 
    HR aligns better with user expectations than precision does, 
    providing a more user-centric evaluation of the system's performance.
It is important to note that HR is undefined for Zero-Result queries 
    due to division by zero. 
Consequently, for this class of queries, 
    we study hallucinations by examining the number of false positives directly.
    

\subsubsection{Hallucination Evaluation}
Figure~\ref{fig:hallucination-ratio} compares HR of ada+GPT4 and Hypercube 
    on the three query categories. 
Hypercube has a much lower tendency to hallucinate than ada+GPT4.
We found that spikes in HR help us identify challenging queries, 
    which we can later study to understand the reasons behind \textit{excessive hallucinations}.
For example, the query 
    ``Early stage breast cancer patients treated surgically other than mastectomies'' 
    led to excessive hallucinations in ada+GPT4. 
    This occurred due to the categorical exclusion of mastectomies and their subtypes 
        from all breast cancer surgeries. 
    Additionally, the vagueness of the phrase ``Early stage'' resulted in over-retrieving incorrect chunks. 
    Furthermore, since a cancer stage can change over time, 
        a patient record could contain other chunks indicating a current \textit{advanced} stage. 
        If the retriever is likely to miss such chunks, the reader will not examine it,
         which highlights the importance of a comprehensive offline longitudinal reasoning.             
Similarly, Hypercube excessively hallucinated on the query ``FIND ME PATIENTS WHO HAVE DIED''. 
    Upon inspection of its knowledge base, we found that it has incorrectly over-extracted death events from patient records.
These findings should influence future iterations in the debugging and development of these ACR systems. 

    Figure \ref{fig:hallucination} compares the false positive (FP) count between ada+GPT4 and Hypercube 
        for Zero-Result queries. 
In only two out of the 42 queries, 
    Hypercube hallucinates just one FP each time.
In contrast, ada+GPT4 generate numerous false positives, 
    a problem that is likely to linearly worsen as the corpus size increases.
Similar to HR, the analysis of the spikes in FP count leads to identifying queries with challenging expressions and combination of criteria. 

Overall, we see that both Hypercube and ada+GPT4 have tendencies to hallucinate,
    but ada+GPT4 hallucinates more consistently and more excessively.
We hypothesize that the symbolic components of Hypercube may act as guardrails 
    and reduces its tendency to hallucinate 
        compared to a generative AI systems like ada+GPT4.


\subsection{Set-theoretic Consistency} 
 
In this section, we elaborate on our discussion in \S\ref{sec:eval-phenomena} 
    and evaluate the consistency of ada+GPT4 and Hypercube. 
We differentiate set-theoretic inconsistencies from hallucinations for two key reasons. 
    First, identifying these inconsistencies involves 
        comparing the system-retrieved cohorts of two different queries 
        rather than comparing a retrieved cohort against a gold standard cohort for the same query. 
    Second, since detecting inconsistencies does not require gold patient-query pairs, 
        we cannot determine whether these inconsistencies are false positives, and thus, hallucinations.
Tables \ref{tab:synonym_query_examples}, \ref{tab:intersection_examples}, and \ref{tab:set-parent_child_evaluation} show the inconsistencies of ada+GPT4 
    on paraphrased, intersection, and subtype queries,
    which we will discuss in details in the following subsections.

\subsubsection{Paraphrased Queries}

\begin{table}
\centering
\begin{tabular}{c|p{2in}|c|p{2in}|cc}
\toprule
$\text{ID}_A$ & Query A  & $\text{ID}_B$ & Query B & $|C_B - C_A|$ & $|C_A - C_B|$ \\
\midrule
\multirow{2}{*}{97} & Find me patients treated with osimertinib  & \multirow{2}{*}{101} & Find me patients treated with Tagrisso & \multirow{2}{*}{3 (23\% of $B$)}  & \multirow{2}{*}{2 (17\% of $A$)} \\
\bottomrule
\end{tabular}
\caption{Example of ada+GPT4 inconsistency on Paraphrased Queries}
\label{tab:synonym_query_examples}
\end{table}

In \S\ref{sec:eval-phenomena}, we defined paraphrased queries as 
    any set of queries that seek the same cohort regardless of their lexical choice. 
For any two paraphrased queries \textit{A} and \textit{B} 
    with two system-retrieved cohorts $C_A$ and $C_B$, respectively, 
    we expect a consistent system to ensure that $C_A = C_B$.
To measure a system's \textit{inconsistency} on such queries,
    we calculate the size of difference between the two cohorts in both directions:
        $|C_A - C_B|$ and $|C_B - C_A|$.

Table~\ref{tab:synonym_query_examples} reports ada+GPT4's paraphrasing inconsistency on two queries
    that refer to patients who received the same medication 
        using the generic name (\textit{osimertinib}) in one and the brand name (\textit{Tagrisso}) in the other. 
ada+GPT4 retrieves 13 patients for the Tagrisso query, 
    3 of whom (23\%) do not appear in the osimertinib query. 
Similarly, it produces 12 patients for osimertinib, 2 of whom (17\%) is not retrieved for Tagrisso.
This inconsistency underscored by the fact that 
    when GPT-4 is asked to generate paraphrases for Tagrisso, it produces osimertinib, and vice versa. 
This indicates that while LLMs store substantial knowledge during training, 
    they still fail to apply that knowledge consistently. 
Moreover, these 5 patients turned out to be false positives, 
    indicating that this inconsistency is not due to a failure of the retriever but of the reader,
        which, despite being resource-intensive, failed to use its knowledge consistently.

Since Hypercube utilizes medical ontologies as part of its large-scale reasoning, 
    it exhibited no inconsistencies for these two queries. 
However, if the paraphrases are not properly mapped to the ontology, 
    remaining consistent on paraphrased queries becomes more challenging. 
Nonetheless, the symbolic components of Hypercube ensure that once it has the relevant knowledge, it will apply it consistently.

\subsubsection{Intersection Queries}
\begin{table}
    \centering
   \begin{tabular}{l|p{2in}|l|p{2.25in}|rrr}
\toprule
ID  & Query A & ID & Query B & $|C_A|$ & $|C_B|$ & $|C_B - C_A|$ \\
\midrule
\multirow{8}{*}{58} &\multirow{8}{*}{Find me patients with  breast cancer} & 113 & breast cancer patients with carcinoma & 613 & 547 & 22 (4\% of $B$) \\
        &       & \multirow{2}{*}{107} & breast cancer patients who received a breast cancer chemo except tamoxifin & \multirow{2}{*}{613} & \multirow{2}{*}{196} & \multirow{2}{*}{10 (5\% of $B$)} \\
        &       & \multirow{2}{*}{105} & Early stage breast cancer patients treated surgically other than mastectomies.& \multirow{2}{*}{613} & \multirow{2}{*}{298} & \multirow{2}{*}{13 (4\% of $B$)} \\
        &       & \multirow{3}{*}{18} & Show me patients with advanced breast cancer and one or more PIK3CA/AKT1/PTEN alterations & \multirow{3}{*}{613} & \multirow{3}{*}{6} & \multirow{3}{*}{2 (33\% of $B$)} \\
\bottomrule
\end{tabular}
    \caption{Examples of set intersection inconsistency of ada+GPT4.}
    \label{tab:intersection_examples}
\end{table}

In \S\ref{sec:eval-phenomena}, we defined complex queries as 
    any query that uses multiple criteria joined with set operations.
In this section, we investigate system's consistency on \textbf{intersection queries},
    a type of complex queries constructed with the AND logical operator. 
Table~\ref{tab:intersection_examples} shows ada+GPT4's inconsistency 
    on a \textit{base} query (Query \textit{A}), ``find me patients with breast cancer,'' 
    and four complex queries (column Query \textit{B}) 
        that narrow the cohort down by specifying further restrictions. 
These restrictive criteria are diverse,
    encompassing aspects such as histology, chemotherapy, surgery, and genetic mutations.
For a system to be consistent, all four intersection queries must return a sub-cohort of the base cohort;
    i.e., $|C_B - C_A| = 0$.
The table shows that ada+GPT4 is inconsistent on all four queries, 
    with \textasciitilde$4$\% (up to 33\% on the most restrictive query) of their patients not appearing in the base query. 
It is important to note that the base query in this example, ``breast cancer,'' 
    is a \textbf{broad query} with approximately 40\% of our corpus included in its gold cohort. 
The narrower the base query, 
    the more challenging it is for an LLM-only system 
        to maintain consistency on more specific queries 
            such as intersection or subtype as the next section will show.

Hypercube's run-time large-scale reasoning
    breaks a query into its sub-queries, executes them independently, 
        and combines their results using set operations such as AND, OR, NOT, and EXCEPT (for set difference). 
    This setup ensures that Hypercube is always consistent on complex queries. 
We observed no inconsistencies for Hypercube on any of the four query pairs in Table~~\ref{tab:intersection_examples}.

\subsubsection{Subtype Queries}
\begin{table*}[h]
    \centering
    \begin{tabular}{c|p{2in}|c|p{2in}|rrr}
\toprule
$\text{ID}_P$ & Parent Query $P$ &$\text{ID}_C$&Child Query $C$& $|C_P|$ & $|C_C|$ & $|C_C - C_P|$ \\
\midrule
\multirow{2}{*}{102} & Find me patients receiving systemic therapy & \multirow{2}{*}{103} &Find me patients receiving targeted therapy & \multirow{2}{*}{651} & \multirow{2}{*}{181} & \multirow{2}{*}{36 (20\% of $C$)} \\
\multirow{2}{*}{103} & Find me patients receiving targeted therapy & \multirow{2}{*}{99}&\multirow{2}{*}{Find me patients treated with a TKI} & \multirow{2}{*}{181} & \multirow{2}{*}{64} & \multirow{2}{*}{24 (38\% of $C$)} \\
\multirow{2}{*}{99} & \multirow{2}{*}{Find me patients treated with a TKI} & \multirow{2}{*}{98} & Find me patients treated with an EGFR TKI & \multirow{2}{*}{64} & \multirow{2}{*}{33} & \multirow{2}{*}{12 (36\% of $C$)} \\
\multirow{2}{*}{98} & Find me patients treated with an EGFR TKI & \multirow{2}{*}{101} & Find me patients treated with Tagrisso & \multirow{2}{*}{33} & \multirow{2}{*}{13} & \multirow{2}{*}{1 (8\% of $C$)} \\
\midrule
\multirow{2}{*}{101} & Find me patients treated with Tagrisso & \multirow{2}{*}{108} & Tagrisso and any other EGFR inhibitor & \multirow{2}{*}{13} & \multirow{2}{*}{25} & \multirow{2}{*}{18 (72\% of $C$)} \\
\bottomrule
\end{tabular}
    \caption{Subtype inconsistency for ada+GPT4}
    \label{tab:set-parent_child_evaluation}
\end{table*}
In \S\ref{sec:eval-phenomena}, we defined \textbf{subtype queries} as 
    any pair of queries consisting of a parent (\textit{P}) and a child (\textit{C}),
        where the child query includes all the criteria of the parent query but specifies at least one criterion further.
Similar to intersection queries above,
    a system's answer to the child query must be a subset of its answer to the parent query.
%
%
%

The first four rows in Table~\ref{tab:set-parent_child_evaluation} present four parent-child pairs
    that are part of a chain of subtype queries getting narrower as you go down the list.
Table~\ref{tab:set-parent_child_evaluation} shows the size of the parent's cohort ($C_P$) and child's cohort ($C_C$) 
    along with ada+GPT4 inconsistencies measured as $|C_C - C_P|$.
ada+GPT4 is inconsistent on all four pairs, 
    retrieving up to 38\% of the child's cohort outside of the parent's cohort. 

The last row of the Table~\ref{tab:set-parent_child_evaluation} 
    shows the pair ``Tagrisso'' and ``Tagrisso and any other EGFR inhibitor.''
Although the latter is technically an intersection query, 
    it can also be considered a subtype query since Tagrisso is an EGFR inhibitor. 
The complexity of this query confuses ada+GPT4, 
    resulting in the child's cohort being 92\% larger than the parent's, 
        with 72\% of the child's cohort not appearing in the parent's.

Hypercube exhibits no inconsistencies across all five pairs of queries, including the last row. 
This consistency is achieved through its large-scale deductive reasoning over medical hierarchies 
    combined with set operations. 
Maintaining this consistency, however, can be challenging 
    if the hierarchies lack essential knowledge artifacts needed to comprehend a child or parent query. 
Once the relevant knowledge is acquired, 
    Hypercube ensures its consistent application throughout the system and across queries.

\subsubsection{Conclusion of Set-Theoretic Consistency}

Unlike retrieval quality and hallucination tendencies, which require ground-truth patient-query pairs for evaluation, 
    consistency evaluation only requires ground-truth pairings of queries based on the types discussed. 
These gold query pairs are much easier and cheaper to construct, 
    enabling the creation of an extensive gold set of such query pairs. 
Additionally, this gold set is independent of the corpus, 
    allowing for consistency evaluations across different corpora from various hospitals. 
This approach helps in observing system consistency and identifying challenging cases in new corpora.

Furthermore, a metric can be developed from these inconsistency measures, 
    providing a holistic evaluation and enabling system comparison. 
Given a gold set and a metric, models can be fine-tuned to enhance their set-theoretic consistency on any given corpus.

\section{Conclusion}

We introduce Automatic Cohort Retrieval (ACR), 
    a crucial task for harnessing the power of real-world data for clinical research. 
We underscore the importance of developing accurate, yet efficient, ACR systems 
    to advance clinical research applications. 
Our contributions include defining the ACR task, 
    creating a publicly available query dataset 
    and an EMR dataset labeled for these queries. 
We introduced 
    an \textit{AI-powered} method for generating large query-patient pair gold datasets.
%
Additionally, we developed a comprehensive evaluation framework with metrics to assess quality, hallucinations, and inconsistencies. 
Our extensive evaluation of three ACR baselines, 
    including LLM-only and neuro-symbolic approaches, 
    highlighted significant weakness areas in retrieval quality 
        as well as improvement opportunities 
    while examining the effectiveness-efficiency trade-offs.

The evaluation demonstrated that while Hypercube currently leads in performance, 
    LLMs exhibit potential in helping automate the retrieval of patient cohorts 
        from extensive, longitudinal datasets. 
Although Hypercube is a commercial black box, 
    the fact that a neuro-symbolic approach can outperform GPT4 
        highlights the potential of integrating expert knowledge with LLMs in healthcare, 
        a domain rich with explicit knowledge.
Furthermore, neuro-symbolic approaches 
    could enhance the practical adoption of ACR systems in the real world
    since healthcare professional often require system control, reduced hallucinations, and consistent, predictable behavior.

Looking ahead, it is imperative that AI researchers
    continues to refine and adapt LLM technologies 
        to meet the specific needs of medical researchers. 
This includes improving model interpretability and controlling hallucinations. 
By addressing these challenges, we can advance the field of ACR, 
    ultimately supporting better clinical decision-making and patient outcomes, 
    and contributing to the development of new treatments and interventions.

\newpage
\bibliographystyle{abbrvnat}
\bibliography{mendel_arxiv}


\newpage
\appendix

\section{Patient Set Statistics}
\label{apd:dataset-stats}
\begin{table}[h]
\centering
\begin{tabular}{llr}
\toprule
Category & Subcategory & Overall \\
\midrule
Age at diagnosis, years &  & 64 $\pm$ 14 \\ \midrule
Sex & Male & 488 \\
    & Female & 929 \\
    & Unknown & 19 \\ \midrule
Race & White & 592 \\
 & Black or African American & 94 \\
 & American Indian or Alaska Native & 0 \\
 & Asian & 241 \\
 & Native Hawaiian or Pacific Islander & 2 \\
 & Other/Unknown & 507 \\ \midrule
Ethnicity & Hispanic or Latino & 58 \\
 & Not Hispanic or Latino & 3 \\
 & Other & 627 \\
 & Unknown & 668 \\ \midrule
Performance Status & ECOG 0 & 380 \\
 & ECOG 1 & 297 \\
 & ECOG 2 & 144 \\
 & ECOG 3 & 49 \\
 & ECOG 4 & 8 \\ \midrule
Stage Group & Stage I & 276 \\ 
 & Stage II & 111 \\
 & Stage III & 146 \\
 & Stage IV & 234 \\ \midrule
Brain metastasis & Yes & 58 \\
 & No & 1378 \\ \midrule
Tumor type & Breast & 568 \\
 & Lung & 176 \\
 & Prostate & 145 \\
 & Colorectal & 33 \\
 & Melanoma & 17 \\
 & Bladder & 50 \\
 & Non-Hodgkin Lymphoma & 0 \\
 & Kidney and renal pelvis & 21 \\
 & Uterus & 0 \\
 & Leukemia & 25 \\
 & Pancreas & 0 \\
 & Thyroid & 36 \\
 & Liver and intrahepatic bile duct & 25 \\
 & Other & 340 \\
\bottomrule
\end{tabular}
\caption{Summary statistics and distributions of patient conditions}
\label{tab:patient_stats}
\end{table}
\begin{figure}[H]
 \centering
  \includegraphics[width=0.5\textwidth]{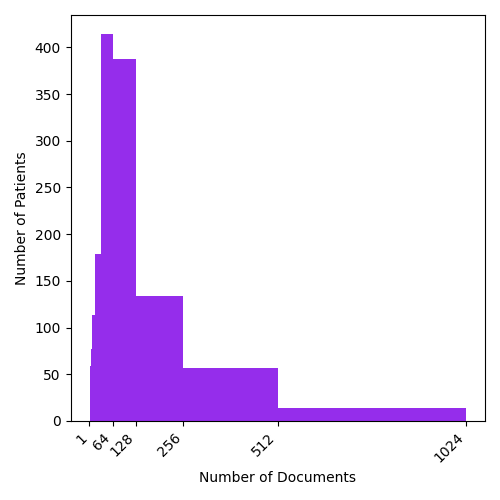}
  \caption{\textbf{Distribution of patients by the quantity of documents.} Documents are not distributed equally, and there is large variance over the amount of information available, that needs to be processed, per patient.}
  \label{fig:num_docs}
\end{figure}


\newpage
\section{Retrieval Results}
\label{apd:results}
\begin{table}[H]
\centering
\begin{footnotesize}
\begin{tabular}{lP{1.75cm}P{1.75cm}P{1.75cm}P{1.75cm}}
\toprule
Model & Precision$\uparrow$ &  Recall$\uparrow$ &  F1$\uparrow$ &  FPR$\downarrow$ \\
\midrule
\multicolumn{5}{c}{\textbf{Broad queries} with gold cohort of $\geq 50$ patients}\\
\midrule
ada & $61.59 \pm 29.5$ & $56.71 \pm 18.24$ & $52.07 \pm 15.94$ & $8.2 \pm 7.41$ \\
v3-large & $67.93 \pm 29.36$ & $50.41 \pm 21.75$ & $49.75 \pm 17.59$ & $5.62 \pm 6.21$ \\
SBERT & $50.18 \pm 27.82$ & $50.96 \pm 17.19$ & $44.85 \pm 17.09$ & $14.46 \pm 10.4$\\ 
$\text{PubMed}_{\text{QA}}$ & $40.05 \pm 16.39$ & $42.28 \pm 15.88$ & $7.31 \pm 5.82$ & $60.02 \pm 30.13$ \\
$\text{BioBERT}_{\text{QA}}$ & $29.33 \pm 25.59$ & $45.05 \pm 7.21$ & $29.63 \pm 17.32$ & $32.71 \pm 7.78$ \\
$\text{PubMed}_{\text{SNLI}}$& $54.74 \pm 31.67$ & $52.88 \pm 13.72$ & $46.49 \pm 16.73$ & $10.73 \pm 8.52$ \\
$\text{BioBERT}_{\text{SNLI}}$ & $49.05 \pm 28.63$ & $48.95 \pm 11.87$ & $42.58 \pm 14.24$ & $12.02 \pm 6.65$\\
ada+GPT4 & $79.38 \pm 21.9$ & $47.09 \pm 25.79$ & $52.68 \pm 20.05$ & $1.87 \pm 1.92$ \\
Hypercube & $76.19 \pm 20.61$ & $87.43 \pm 8.67$ & $79.4 \pm 12.82$ & $12.81 \pm 22.53$ \\
\midrule
\multicolumn{5}{c}{\textbf{Narrow queries} with gold cohort of  $\geq 10$ patients and $< 50$ patients}\\
\midrule
ada & $10.68 \pm 9.66$ & $68.29 \pm 19.08$ & $16.84 \pm 12.02$ & $12.17 \pm 5.16$\\
v3-large & $14.67 \pm 11.14$ & $62.63 \pm 20.49$ & $21.08 \pm 11.33$ & $8.33 \pm 5.64$\\
SBERT & $6.02 \pm 4.8$ & $68.08 \pm 21.1$ & $10.7 \pm 7.93$ & $21.89 \pm 8.76$\\ 
$\text{PubMed}_{\text{QA}}$ & $61.61 \pm 25.92$ & $17.54 \pm 11.49$ & $9.8 \pm 3.91$ & $11.06 \pm 9.16$\\
$\text{BioBERT}_{\text{QA}}$ & $2.53 \pm 1.8$ & $49.19 \pm 14.27$ & $4.67 \pm 3.13$ & $35.44 \pm 5.35$ \\
$\text{PubMed}_{\text{SNLI}}$ & $8.28 \pm 7.95$ & $73.81 \pm 17.54$ & $13.71 \pm 10.61$ & $17.06 \pm 5.49$\\
$\text{BioBERT}_{\text{SNLI}}$ & $6.02 \pm 3.11$ & $67.52 \pm 18.65$ & $10.81 \pm 5.19$ & $18.55 \pm 5.19$\\
ada+GPT4 & $38.34 \pm 18.3$ & $50.68 \pm 21.87$ & $41.23 \pm 16.55$ & $1.83 \pm 2.35$\\
Hypercube & $56.1 \pm 21.85$ & $54.83 \pm 24.91$ & $52.02 \pm 17.66$ & $1.06 \pm 1.31$ \\
\midrule
\multicolumn{5}{c}{\textbf{Sparse queries} with gold cohort of 1 to 9 patients}\\
\midrule
ada & $1.14 \pm 2.33$ & $78.95 \pm 64.95$ & $2.25 \pm 4.5$ & $13.57 \pm 5.99$ \\
v3-large& $2.23 \pm 2.6$ & $84.21 \pm 67.84$ & $4.34 \pm 5.01$ & $7.35 \pm 5.58$ \\
SBERT & $0.72 \pm 1.79$ & $75.44 \pm 62.38$ & $1.42 \pm 3.48$ & $20.73 \pm 7.64$\\ 
$\text{PubMed}_{\text{QA}}$ & $1.17 \pm 3.15$ & $52.63 \pm 49.65$ & $2.29 \pm 5.92$ & $1.23 \pm 1.45$\\
$\text{BioBERT}_{\text{QA}}$ & $0.37 \pm 3.93$ & $67.54 \pm 65.11$ & $0.74 \pm 7.41$ & $35.9 \pm 3.63$\\
$\text{PubMed}_{\text{SNLI}}$ & $0.85 \pm 2.07$ & $83.33 \pm 64.16$ & $1.68 \pm 4$ & $19.4 \pm 6.43$ \\
$\text{BioBERT}_{\text{SNLI}}$ & $0.81 \pm 2.57$ & $70.18 \pm 53.31$ & $1.6 \pm 4.9$ & $17.1 \pm 4.73$ \\
ada+GPT4 & $17.87 \pm 21.81$ & $50 \pm 51.3$ & $26.33 \pm 30.6$ & $0.46 \pm 0.51$ \\
Hypercube & $32.64 \pm 28.32$ & $41.23 \pm 57.16$ & $36.43 \pm 37.87$ & $0.17 \pm 0.45$\\
\midrule
\multicolumn{5}{c}{\textbf{Zero-Result queries} with empty gold cohorts}\\
\midrule
ada &  &  &  & $12.67 \pm 5.99$\\
v3-large&  &  &  & $8.51 \pm 5.55$ \\
SBERT &  &  &  & $20.29 \pm 6.99$ \\ 
$\text{PubMed}_{\text{QA}}$ & &  &  & $2.38 \pm 15.43$ \\
$\text{BioBERT}_{\text{QA}}$ &  &  &  & $36.67 \pm 6.27$ \\
$\text{PubMed}_{\text{SNLI}}$ &  &  &  & $19.44 \pm 7.49$ \\
$\text{BioBERT}_{\text{SNLI}}$ &  &  &  & $17.36 \pm 5.89$ \\
ada+GPT4 &  &  &  & $0.21 \pm 0.27$ \\
Hypercube &  &  &  & $0.06 \pm 0.38$ \\
\bottomrule
\end{tabular}
\end{footnotesize}
\caption{Full Cohort Retrieval Results}
\label{tab:full-cohort-retrieval-results}
\end{table}

\newpage
\section{Evaluation Framework Detailed Description}
\label{apd:checklist}

\begin{itemize}
    \item Retrieve cohorts based on simple tumor and treatment descriptors at various levels of abstraction
    \item Retrieve cohorts based on complex tumor descriptors drawn from OncoTree \citep{2021_oncotree}, reflecting precise categorizations not present in standard biomedical terminologies.
    \item Retrieve cohorts eligible or ineligible for clinical research, based largely on the various arms of the phase II basket study TAPUR
    \item Retrieve cohorts of patients eligible for newly approved indications for targeted therapies, reflecting the latest advances in precision medicine
    \item Retrieve cohorts of patients who have been tested for somatic or germline alterations relevant to targeted therapies as of June 2021, reflecting a comprehensive set of precision oncology RWE generation and decision-making opportunities
    \item Retrieve cohorts of patients based on different levels of abstraction used in clinical reasoning: categories of medication, semantic ordering among clinical concepts, explicit and implicit exceptions.
\end{itemize}
\newpage

\section{Question Bank}
\label{apd:question-bank}

\begin{table*}[ht]
    \centering
    \small
\begin{tabular}{p{0.5in}|p{5.5in}}
\toprule
Query ID & Question \\ \midrule
1 & Find me prostate cancer patients with a HRR mutation \\ \midrule
2 & Find me HRRm mCRPC patients \\ \midrule
3 & Find me NSCLC patients with an EGFR nonresistant mutation other than exon 19 del and L858R \\ \midrule
4 & Find me patients with bladder cancer and an oncogenic mutation in FGFR3 \\ \midrule
5 & Find me patients with a solid tumor and an NTRK fusion \\ \midrule
6 & Find me patients with a solid tumor that is MSI high \\ \midrule
7 & Find me patients with a solid tumor that has high tumor mutational burden \\ \midrule
8 & breast cancer patients where T stage is between T2b and T3c and N stage is no more than N3b. However, if N stage is N0, then stage must be T3 or higher.  \\ \midrule
9 & CRC patients with Stage no less than IIc treated with folfiri \\ \midrule
10 & ECOG is 2 or less \\ \midrule
11 & histologically confirmed T stage of at least T2c \\ \midrule
12 & find pts with pathological N stage of at least N1 \\ \midrule
13 & Find me patients with ACTIONABLE KRAS MUTATIONS \\ \midrule
14 & Show me patients with VHL and some other cancer \\ \midrule
15 & Show me patients with CLL/SLL and treatment with a BTK inhibitor and a BCL-2 inhibitor \\ \midrule
16 & Show me patients with desmoid tumors \\ \midrule
17 & Show me patients with non-metastatic castration-sensitive prostate cancer (nmCSPC) \\ \midrule
18 & Show me patients with advanced breast cancer and one or more PIK3CA/AKT1/PTEN alterations \\ \midrule
19 & Show me patients with locally advanced or metastatic solid tumor, multiple myeloma or B cell non-Hodgkin lymphoma, ECOG 0-2, no primary brain tumors or leptomeningeal metastases, and one of the following genomic alterations: CDKN2A deletion or mutation, CDK4, CDK6 amplifications, CDKN2B deletion or mutation. \\ \midrule
20 & Show me patients with locally advanced or metastatic solid tumor, multiple myeloma or B cell non-Hodgkin lymphoma, ECOG 0-2, no primary brain tumors or leptomeningeal metastases, and one of the following genomic alterations: CSF1R, PDGFR, VEGFR1/2/3, KIT, FLT-3, RET, FGFR1/2/3, VHL amplifications or mutations  \\ \midrule
21 & Show me patients with locally advanced or metastatic solid tumor, multiple myeloma or B cell non-Hodgkin lymphoma, ECOG 0-2, no primary brain tumors or leptomeningeal metastases, and one of the following genomic alterations: mTOR, TSC1/2, AKT1 mutations \\ \midrule
22 & Show me patients with locally advanced or metastatic solid tumor, multiple myeloma or B cell non-Hodgkin lymphoma, ECOG 0-2, no primary brain tumors or leptomeningeal metastases, and one of the following genomic alterations: ERBB2 amplification or overexpression, and specific ERBB2 mutations. \\ \midrule
23 & Show me patients with locally advanced or metastatic solid tumor, multiple myeloma or B cell non-Hodgkin lymphoma, ECOG 0-2, no primary brain tumors or leptomeningeal metastases, and one of the following genomic alterations: BRAF V600E/D/K/R mutations \\ \midrule
24 & Show me patients with locally advanced or metastatic solid tumor, multiple myeloma or B cell non-Hodgkin lymphoma, ECOG 0-2, no primary brain tumors or leptomeningeal metastases, and one of the following genomic alterations: RET, VEGFR1/2/3, KIT, PDGFRG, RAF-1, BRAF mutations or amplifications \\
\bottomrule
\end{tabular}
\end{table*}

\begin{table*}[ht]
    \centering
    \small
\begin{tabular}{p{0.5in}|p{5.5in}}
\toprule
Query ID & Question \\ \midrule
25 & Show me patients with locally advanced or metastatic solid tumor, multiple myeloma or B cell non-Hodgkin lymphoma, ECOG 0-2, no primary brain tumors or leptomeningeal metastases, and one of the following genomic alterations: germline or somatic BRCA1/2 inactivating mutations; ATM mutations or deletions \\ \midrule
26 & Show me patients with locally advanced or metastatic solid tumor, multiple myeloma or B cell non-Hodgkin lymphoma, ECOG 0-2, no primary brain tumors or leptomeningeal metastases, and one of the following genomic alterations: specific POLE and POLD1 mutations  \\ \midrule
27 & Show me patients with locally advanced or metastatic solid tumor, multiple myeloma or B cell non-Hodgkin lymphoma, ECOG 0-2, no primary brain tumors or leptomeningeal metastases and one of the following genomic alterations: MSI high status, high tumor mutational burden, MLH1, MSH2/6, PMS2, EPCAM mutations, specific POLE or POLD1 mutations, BRCA1/2, ATM, MSH3, PMS1, MLH3, EXO1, RFC1/2/3/4/5, PCNA, RPA1/2/3/4, and SSBP1 loss of function mutations. \\ \midrule
28 & Show me patients with locally advanced or metastatic solid tumor, multiple myeloma or B cell non-Hodgkin lymphoma, ECOG 0-2, no primary brain tumors or leptomeningeal metastases and one of the following genomic alterations: germline or somatic BRCA1/2 and PALB2 mutations \\ \midrule
29 & Show me patients with locally advanced or metastatic solid tumor, multiple myeloma or B cell non-Hodgkin lymphoma, ECOG 0-2, no primary brain tumors or leptomeningeal metastases and one of the following genomic alterations: ERBB2 amplification or overexpression \\ \midrule
30 & Show me patients with locally advanced or metastatic solid tumor, multiple myeloma or B cell non-Hodgkin lymphoma, ECOG 0-2, no primary brain tumors or leptomeningeal metastases and one of the following genomic alterations: germline or somatic mutations in BRCA1/2, PALB2, ATM, ATR, CHEK2, FANCA, RAD51C, NBN, MLH1, MRE11A, CDK12; positive genomic instability score reported on the Myriad MyChoice CDx test; or Genomic Loss of Heterozygosity (LOH) Score above threshold \\ \midrule
31 & Show me patients with locally advanced or metastatic solid tumor, multiple myeloma or B cell non-Hodgkin lymphoma, ECOG 0-2, no primary brain tumors or leptomeningeal metastases and one of the following genomic alterations: ROS1 fusion \\ \midrule
32 & Show me patients with locally advanced or metastatic solid tumor, multiple myeloma or B cell non-Hodgkin lymphoma, ECOG 0-2, no primary brain tumors or leptomeningeal metastases and one of the following genomic alterations: NTRK1/2/3 amplification \\ \midrule
34 & Show me patients with locally advanced or metastatic solid tumor, multiple myeloma or B cell non-Hodgkin lymphoma, ECOG 0-2, no primary brain tumors or leptomeningeal metastases and one of the following genomic alterations: FGFR 1,2,3,4 fusion (or other rearrangement) or mutation \\ \midrule
35 & Find me patients with ductal carcinoma in situ \\ \midrule
36 & Find me patients with ductal carcinoma in situ of the breast \\ \midrule
37 & Find me patients with metastatic castration resistant prostate cancer \\ \midrule
38 & Find me patients with high grade serous ovarian cancer \\ \midrule
39 & Find me patients with advanced NSCLC \\ \midrule
40 & Find me patients with lung adenocarcinoma \\ \midrule
41 & Find me patients with pleuropulmonary blastoma \\ \midrule
42 & Find me patients with metaplastic breast cancer \\ \midrule
43 & Find me patients with clear cell sarcoma \\ \midrule
44 & Find me patients with synovial sarcoma \\ \midrule
45 & Find me patients with Ewing sarcoma \\ \midrule
46 & Find me patients with leiomyosarcoma \\
\bottomrule
\end{tabular}
\end{table*}

\begin{table*}[ht]
    \centering
    \small
\begin{tabular}{p{0.5in}|p{5.5in}}
\toprule
Query ID & Question \\ \midrule
47 & Find me patients with lung neuroendocrine tumor \\ \midrule
48 & Find me patients with sarcomatoid carcinoma of the lung \\ \midrule
49 & Find me patients with pleomorphic carcinoma of the lung \\ \midrule
50 & Find me patients with endometroid ovarian cancer \\ \midrule
51 & Find me patients with head and neck squamous cell carcinoma \\ \midrule
52 & Find me patients with adenomyoepithelioma of the breast \\ \midrule
53 & Find me patients with solid papillary carcinoma of the breast \\ \midrule
54 & Find me patients with spindle cell carcinoma of the lung \\ \midrule
55 & Find me patients with clear cell ovarian cancer \\ \midrule
56 & Find me patients with low grade serous ovarian cancer \\ \midrule
57 & Find me patients with lung cancer \\ \midrule
58 & Find me patients with breast cancer \\ \midrule
59 & Find me patients with ovarian cancer \\ \midrule
60 & Find me patients with prostate cancer \\ \midrule
61 & Find me patients with soft tissue sarcoma \\ \midrule
62 & Find me patients with kidney cancer \\ \midrule
63 & Find me patients with renal cell carcinoma \\ \midrule
64 & Find me patients with renal clear cell carcinoma \\ \midrule
65 & Find me patients with Wilms' tumor \\ \midrule
66 & Find me NSCLC patients with an ALK fusion \\ \midrule
67 & Find me melanoma patients with BRAF V600E mutation \\ \midrule
68 & Find me NSCLC patients with a BRAF V600E mutation \\ \midrule
69 & Find me CRC patients with a BRAF V600E mutation \\ \midrule
70 & Find me melanoma patients with a BRAF V600K mutation \\ \midrule
71 & Find me patients with a deleterious or suspected deleterious mutation in BRCA1 and/or BRCA2 with ovarian cancer, falliopian tube cancer, pancreatic adenocarcinoma, Her2- breast cancer, or peritoneal cancer \\ \midrule
72 & Find me prostate cancer patients with a deleterious or suspected deleterious mutation in ATM, BARD1, BRIP1,
CDK12, CHEK1, CHEK2, FANCL, PALB2,
RAD51B, RAD51C, RAD51D, and RAD54L \\ \midrule
73 & Find me NSCLC patients with an exon 19 deletion in EGFR \\ \midrule
74 & Find me NSCLC patients with an L858R mutation in EGFR \\ \midrule
75 & Find me NSCLC patients with an EGFR exon 20 insertion \\ \midrule
76 & Find me NSCLC patients with an EGFR T790M mutation \\ \midrule
77 & Find me patients with breast cancer, esophagogastric cancer, gastric cancer, or gastroesophageal junction cancer, and ERBB2 amplification \\ \midrule
78 & Find me patients with bladder cancer or cholangiocarcinoma patients with a FGFR2 fusion \\ \midrule
79 & Find me patients with ovarian cancer who are GIS positive or HRD positive \\
\bottomrule
\end{tabular}
\end{table*}

\begin{table*}[ht]
    \centering
    \small
\begin{tabular}{p{0.5in}|p{5.5in}}
\toprule
Query ID & Question \\ \midrule
81 & Find me NSCLC patients with KRAS G12C mutation \\ \midrule
82 & Find me NSCLC patients with a MET exon 14 skipping mutation \\ \midrule
83 & Find me patients with CRC and dMMR or MSI-H \\ \midrule
84 & Find me patients with endometrial cancer and dMMR or MSI-H \\ \midrule
85 & Find me patients with PDGFRA exon 18 mutation and GIST  \\ \midrule
86 & Find me patients with HR+/Her2- breast cancer and a mutation in PIK3CA \\ \midrule
87 & Find me patients with NSCLC or thyroid cancer and a RET fusion \\ \midrule
88 & Find me patients with medullary thyroid cancer and an oncogenic mutation in RET \\ \midrule
89 & Find me patients with NSCLC and a ROS1 fusion \\ \midrule
90 & Find me patients with an oncogenic mutation in NF1 \\ \midrule
91 & Find me patients with COL1IA1-PDGFB fusions \\ \midrule
92 & Find me patients with SMARCB1 deletions \\ \midrule
93 & Find me patients with oncogenic mutations TSC1 and TSC2 \\ \midrule
94 & Find me patients with KIT exon 11, 9, 13, 14, and 17 mutations \\ \midrule
95 & Find me CRC patients with KRAS and/or NRAS exon 2, 3, and 4 mutations \\ \midrule
97 & Find me patients treated with osimertinib \\ \midrule
98 & Find me patients treated with an EGFR TKI \\ \midrule
99 & Find me patients treated with a TKI  \\ \midrule
100 & Find me patients treated with RxCUI code 1721560 \\ \midrule
101 & Find me patients treated with Tagrisso \\ \midrule
102 & Find me patients receiving systemic therapy \\ \midrule
103 & Find me patients receiving targeted therapy \\ \midrule
104 & NSCLC patients who recevied any systemic therapy except EGFR inhibitors \\ \midrule
105 & Early stage breast cancer patients treated surgically other than mastectomies. \\ \midrule
106 & triple negative ductal or lobular carcinoma except in situ \\ \midrule
107 & breast cancer patients who received a breast cancer chemo except tamoxifin \\ \midrule
108 & Tagrisso and any other EGFR inhibitor  \\ \midrule
109 & ypT3 rectal cancer patients \\ \midrule
110 & ypT3 rectal cancer patients who received neoadjuvant chemoradiation \\ \midrule
111 & breast cancer with T2N0M0 \\ \midrule
112 & patients with biomarkers documented as exactly positive (no strongly or weakly) \\ \midrule
113 & breast cancer patients with carcinoma  \\ \midrule
117 & Find me patients with GERMLINE BUT NOT SOMATIC MUTATIONS IN BRCA or ATM \\ \midrule
114 & find patients with exactly wild type KRAS \\ \midrule
115 & FIND ME PATIENTS WHO HAVE DIED \\ \midrule
118 & Find me patients with a solid tumor and a fusion in NTRK1, NTRK2, or NTRK3 \\
\bottomrule
\end{tabular}
    \caption{Complete List of questions}
    \label{tab:questionbank}
\end{table*}

\end{document}